\documentclass[final,3p,times,twocolumn]{elsarticle}

    \usepackage[utf8]{inputenc}
    \usepackage[english]{babel}
    \usepackage{multirow}
    \usepackage{amsmath, amssymb, amsthm}
    \usepackage{bbm}        
    
    \usepackage{algorithm}
    \usepackage{algpseudocode}
    
    \usepackage{graphicx}
    \usepackage{float}
    \usepackage{booktabs}
    \usepackage{enumitem}
    \usepackage{layout}
    \usepackage{blindtext}
        \usepackage{appendix}

    \usepackage{hyperref}
    \hypersetup{colorlinks=true, linkcolor=black, filecolor=black, urlcolor=cyan}
    
    \usepackage{caption}
    \journal{ICT Express}
    
\begin{document}
    
    \begin{frontmatter}
    
    \title{Graph Attention Networks with Physical Constraints for Anomaly Detection}
    \author[1]{Mohammadhossein Homaei\corref{cor1}}
    \ead{mhomaein@alumnos.unex.es}

    \author[2]{Iman Khazrak}
    \ead{ikhazra@bgsu.edu}
    
    \author[1]{Ruben Molano}
    \ead{rmolano@unex.es}

    \author[1]{Andres Caro}
    \ead{andresc@unex.es}
    
    \author[1]{Mar Avila}
    \ead{mmavila@unex.es}

    \address[1]{Departamento de Ingeniería de Sistemas Informáticos y Telemáticos, Universidad de Extremadura, Av/ Universidad S/N, 10003, Cáceres, Extremadura, Spain}

    \address[2]{Department of Computer Science, Bowling Green State University, Ohio, USA}
    
    \cortext[cor1]{Corresponding author}

    \begin{abstract}
Water distribution systems (WDSs) face increasing cyber-physical risks, which make reliable anomaly detection essential. Many data-driven models ignore network topology and are hard to interpret, while model-based ones depend strongly on parameter accuracy. This work proposes a hydraulic-aware graph attention network using normalized conservation law violations as features. It combines mass and energy balance residuals with graph attention and bidirectional LSTM to learn spatio-temporal patterns. A multi-scale module aggregates detection scores from node to network level. On the BATADAL dataset, it reaches $F1=0.979$, showing $3.3$pp gain and high robustness under $15\%$ parameter noise.  \end{abstract}
    
    
    \end{frontmatter}
    
    
    \section{Introduction}\label{sec:introduction}
    
Water distribution systems (WDSs) are key infrastructures for public health and economy. In recent years, they face more cyber-physical attacks like data manipulation or actuator control~\cite{Taormina2018BATADAL, Ramotsoela2019}. If not detected, such anomalies may cause water loss, service stop, or contamination. SCADA systems collect large sensor data (pressure, flow, tank levels) that support data-based protection~\cite{Homaei2024digital}.

Anomaly detection methods are usually of two kinds. Data-driven ones, such as Deep Learning and Autoencoders~\cite{Taormina2018DeepLearning, Addeen2024}, often ignore hydraulic topology and act as black boxes, limiting trust~\cite{BarredoArrieta2020}. Model-based ones, using simulators like EPANET with residuals~\cite{Housh2018ModelBased}, give physical sense but depend on uncertain parameters and scale poorly. The challenge is to design a model that is both accurate and interpretable.

This work joins both ideas: physics constraints and neural learning. The question is simple:

\begin{quote}
Can a GNN that uses physics violation and topology improve anomaly detection in WDSs, with better robustness and explainability?
\end{quote}

We propose the Physics-GAT model with three parts: (i) physics-informed (PI) features from conservation laws, (ii) Graph Attention Network (GAT) for spatial relations, and (iii) Multi-Scale Detection merging local and global scores. On BATADAL, it gets F1=0.979 vs 0.946 (gain 3.3 points, Wilcoxon $p=0.0009$) and detects 10.6\% faster (TTD: 1.44h vs 1.61h). Ablation shows PI features are most critical.

Section~\ref{sec:related} reviews related works, Section~\ref{sec:methodology} explains the framework, Section~\ref{sec:results} presents experiments, Section~\ref{discussion} discusses findings, and Section~\ref{sec:conclusion} concludes the paper.

\section{Related Work}\label{sec:related}

Early works on WDS anomaly detection used statistical techniques such as CUSUM and EWMA, but these methods had difficulties dealing with the non-linear behavior of SCADA data~\cite{Kanyama2024}. Later, machine learning improved the ability to recognize patterns~\cite{Ramotsoela2019}, while deep models like autoencoders and CVAEs~\cite{Taormina2018DeepLearning, Addeen2024} showed higher accuracy. However, they usually ignore the hydraulic structure of the system and are not easy to interpret~\cite{Sikder2023DeepH2O, BarredoArrieta2020}, learning mostly time dependencies without capturing spatial relations among sensors.

The BATADAL benchmark~\cite{Taormina2018BATADAL, Zischg2017} introduced C-Town as a reference testbed. Its best solution used EPANET simulation combined with residual analysis~\cite{Housh2018ModelBased}, showing the value of physics-based detection. Still, such methods depend on parameters like pipe roughness and water demand, which are often unknown in old networks, reducing their stability and generalization.

More recent graph-based approaches (GCN, GraphSAGE~\cite{Sikder2023DeepH2O}) include the network topology but do not directly use the physical conservation laws. PI-neural networks have achieved good results in other areas such as power grids, HVAC systems, and supply chain resilience~\cite{nejad2025building}, but their use in WDSs is still rare.

Main gaps remain in: (1) dependency on uncertain parameters in model-based methods, (2) deep models that ignore topology, (3) missing explicit use of conservation features, and (4) lack of multi-scale detection for fault localization.

\section{Problem-Driven Methodology}\label{sec:methodology}

\subsection{Problem Formulation and Motivation}\label{sec:problem_formulation}

Detecting anomalies in WDSs is hard since attacks or faults often appear as small and non-stationary changes in normal operation. Data-driven models usually ignore network topology or become unstable under uncertain hydraulic parameters, while model-based ones need fine calibration that is not easy in practice. We model this as a spatio-temporal detection task on a graph $\mathcal{G} = (\mathcal{V}, \mathcal{E})$ with $N$ nodes (junctions, tanks, pumps) and edges as pipes. At each time $t$, SCADA readings form $\mathbf{X}_t \in \mathbb{R}^{N \times F}$ with raw and physics-informed features. The goal is to learn

\[
f: (\mathbf{X}_{t-w+1:t}, \mathcal{G}) \mapsto \mathbf{a}_t \in [0,1]^N,
\]

where $\mathbf{a}_t$ are node-level anomaly scores from a window of length $w$. The mapping must combine temporal context and network structure, respecting physical laws so that anomalies reflect real violations, not model noise.

Our Physics-GAT follows this idea: PI features make the model sensitive to conservation errors, graph attention layers encode spatial relations, and temporal units separate short noise from persistent anomalies.

\subsection{PI Data Representation}\label{sec:pi_representation}

Each node $v_i$ is described by an input vector $\mathbf{h}_t(v_i)$ that concatenates raw SCADA signals ($p_i, Q_i, \ell_i$), temporal statistics and physics-informed (PI) indicators that measure conservation law deviations. PI features rely on available hydraulic quantities (e.g. estimated demand $D_i(t)$, pipe roughness $C_{ij}$), but we use normalization to reduce sensitivity to parameter errors and keep the features informative.

Mass balance violation at node $v_i$ is defined as

\begin{equation}
\phi_{\text{mass}}^i(t) = \frac{\left| \sum_{j \in \mathcal{N}_{\text{in}}(i)} 
Q_{ji}(t) - \sum_{k \in \mathcal{N}_{\text{out}}(i)} Q_{ik}(t) - D_i(t) \right|}
{\sum_{j} Q_{ji}(t) + \epsilon}
\label{eq:mass_violation}
\end{equation}

This relative form reduces the effect of absolute demand or roughness errors by scaling residuals by typical inflow magnitude.

Energy gradient violation between nodes $i$ and $j$ is

\begin{equation}
\phi_{\text{energy}}^{ij}(t) = \frac{| (p_i + z_i) - (p_j + z_j) - h_L(Q_{ij}) |}{\max(p_i + z_i, p_j + z_j)}
\label{eq:energy_violation}
\end{equation}

where $z_i, z_j$ are elevations and $h_L(Q_{ij})$ is the theoretical head loss (Hazen–Williams with known $L_{ij}, D_{ij}, C_{ij}$). The final node vector $\mathbf{h}_t(v_i)$ includes these PI features alongside raw and temporal descriptors.

When nodes are unmeasured, we use pressure-driven interpolation to approximate their feature vectors. For an unmeasured node $v_j$ the representation is

\[
\mathbf{h}_t(v_j) = \sum_{i \in \mathcal{N}_m(j)} w_{ij}\,\mathbf{h}_t(v_i), \quad
w_{ij} = \frac{\exp(-d_{ij}/\sigma)}{\sum_k \exp(-d_{jk}/\sigma)}, \quad \sigma=2,
\]

where $\mathcal{N}_m(j)$ are measured neighbors and $d_{ij}$ is hydraulic distance. This maintains hydraulic continuity and enables full-graph reasoning.

\begin{figure*}
    \centering
    \includegraphics[width=0.9\linewidth]{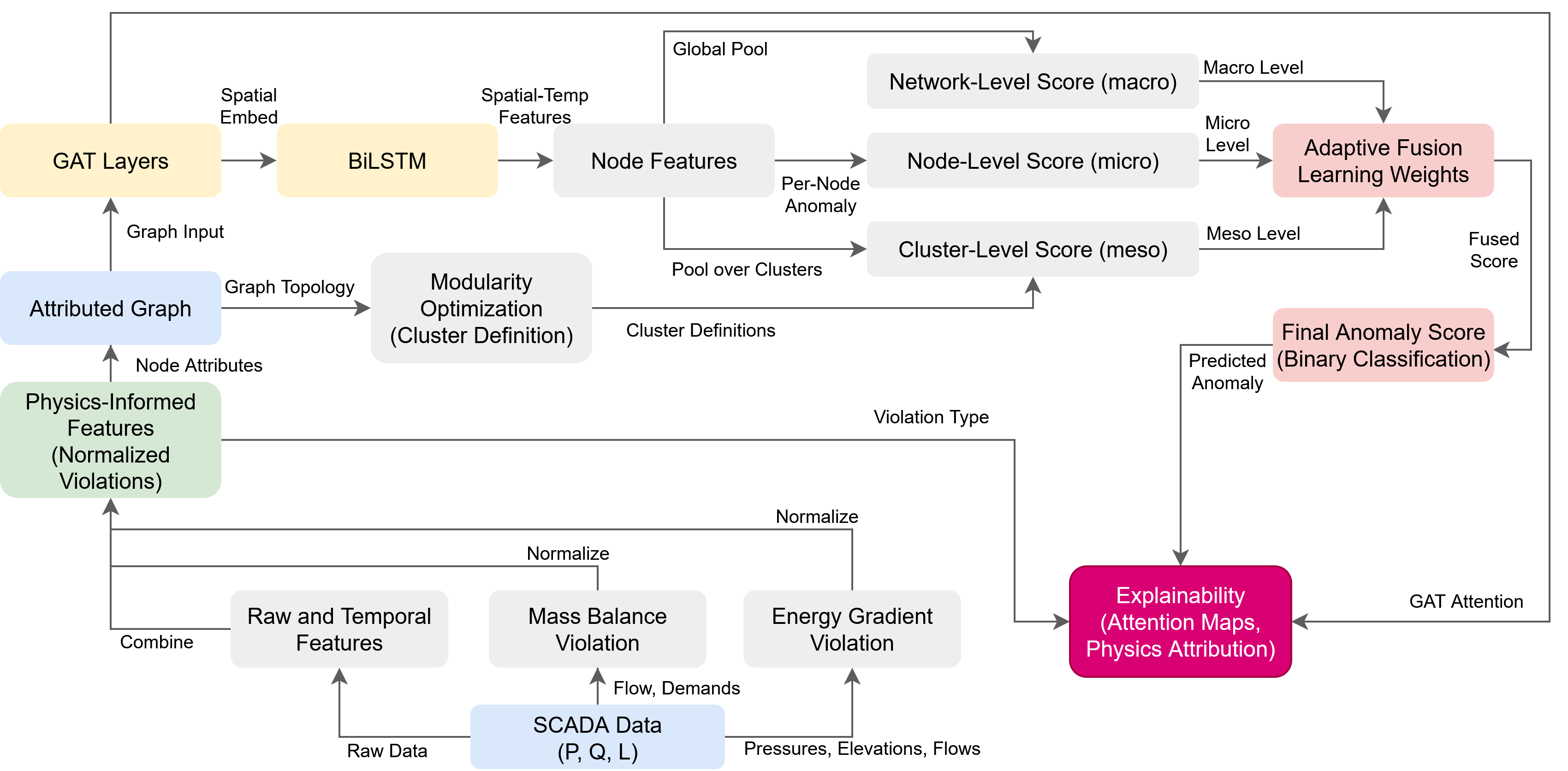}
    \caption{Physics-GAT workflow: physics-informed features feed a GAT-BiLSTM core and an adaptive multi-scale fusion that produces node-level anomaly scores.}
    \label{fig:framework}
\end{figure*}

The overall workflow of the proposed Physics-GAT model, including the integration of physics-informed features and temporal fusion, is illustrated in Fig.~\ref{fig:framework}.

\subsection{Spatio-Temporal Detection Core}\label{sec:core}

The detection core mixes a spatial GAT with a temporal BiLSTM. Three GAT layers ($L=3$) learn adaptive attention $\alpha_{ij}$ showing how anomalies move through the network. Multi-head attention ($K=8$) keeps learning stable. Spatial outputs $\mathcal{Z} = [\mathbf{Z}_{t-w+1},\ldots,\mathbf{Z}_t]$ go into a Bidirectional LSTM with window $w$ (24 hours in tests) to capture both time directions. The fused spatial-temporal features form node vectors $\mathbf{f}_{i,t}$.

To reduce false alarms and make results clearer, scoring is hierarchical. The network is clustered into $\mathcal{C}_k$ using Louvain for hydraulic coherence. We compute node $a_t^{\text{micro}}$, cluster $a_t^{\text{meso}}$ (attention pooling), and global $a_t^{\text{macro}}$ scores. The final node score is:

\begin{equation}
a_t^{\text{final}}(v_i) = \lambda_1 a_t^{\text{micro}}(v_i) + \lambda_2 a_t^{\text{meso}}(\mathcal{C}(i)) + \lambda_3 a_t^{\text{macro}},
\end{equation}

where $\lambda_k$ are softmax weights adapting the scale influence.

\subsection{Training Objective and Explainability}\label{sec:training}

The model is trained with a composite loss

\[
\mathcal{L} = \mathcal{L}_{\text{BCE}} + \lambda_p \mathcal{L}_{\text{physics}} + \lambda_c \mathcal{L}_{\text{consist}},
\]

where $\mathcal{L}_{\text{BCE}}$ is binary cross-entropy for node-level detection, $\mathcal{L}_{\text{physics}}$ is a regularizer that penalizes physics violations during normal operation, and $\mathcal{L}_{\text{consist}}$ enforces spatio-temporal coherence. The physics term is

\begin{equation}
\mathcal{L}_{\text{physics}} = \frac{1}{N} \sum_{i=1}^N \mathbf{1}(y_{i,t}=0) \cdot \max(\phi_{\text{mass}}^i(t), \phi_{\text{energy}}^{i\cdot}(t))
\label{eq:physics_loss}
\end{equation}

and reduces the model tendency to accept unrealistic violations as normal. We optimize with Adam and cosine annealing; in experiments hyperparameters are set to $\lambda_p=0.1$ and $\lambda_c=0.05$ (details and additional tuning are in the Appendix).

Explainability derives from two sources: attention coefficients $\alpha_{ij}$ that indicate likely propagation paths of anomalies, and physics-violation attribution which identifies whether mass or energy inconsistency drives the alarm (via $\arg\max\{\phi_{\text{mass}}^i,\phi_{\text{energy}}^{i\cdot}\}$). These outputs support operator interpretation and post-hoc diagnosis.

\subsection{Benchmark Datasets and Multi-Network Validation}\label{sec:datasets}

We evaluate generalization and scalability using four networks. The BATADAL benchmark on C-Town ($N=128$) provides the main controlled experiments. To test robustness across topologies and sizes we use the 2024 Multi-Network Cyber-Physical Dataset, which includes D-Town ($N\approx 200$) for generalization checks, L-Town ($N>300$) for scalability, and Modena ($N\approx 180$) for realism. This multi-network validation demonstrates that physics-informed, topology-aware learning transfers beyond a single calibrated network and supports practica

 \section{Experimental Results}\label{sec:results}

We test Physics-GAT on the BATADAL benchmark against strong baselines. Evaluation covers global accuracy, ablation, and robustness to real-world noise.

\subsection{Setup and Baselines}

We report F1-score and Time-to-Detection (TTD, hours). Baselines are the model-based BATADAL winner (B1)~\cite{Housh2018ModelBased}, deep models (CVAE~\cite{Addeen2024}, BiLSTM~\cite{Sikder2023DeepH2O}), and graph-free ones (RF, SVM). Physics-GAT runs on PyTorch Geometric with 3 GAT layers (8 heads, 128 hidden units) and a 24-hour input window. Training uses Adam ($\eta_0=10^{-3}$), batch 32, and tuned regularization $\lambda_p=0.1$, $\lambda_c=0.05$. Significance is tested by 1000 bootstrap resamples for 95\% CI and Cohen’s d for effect size.

    \subsection{Overall Performance Comparison}
    
Table~\ref{tab:overall_performance} presents the test results, showing that Physics-GAT clearly outperforms all baselines.    
    \begin{table}[h]
    \scriptsize
    \centering
\caption{Performance on BATADAL test set. 95\% CI obtained via bootstrap (n=1000). Cohen's d vs B1: 1.24.}   
\label{tab:overall_performance}
    \resizebox{\columnwidth}{!}{%
    \begin{tabular}{lccc}
    \toprule
    Method & F1 & CI & TTD (h) \\
    \midrule
    BiLSTM & 0.906 & [.894,.918] & 1.85 \\
    CVAE & 0.938 & [.927,.948] & 1.72 \\
    GCN+BiLSTM & 0.941 & [.931,.951] & 1.69 \\
    GraphSAGE+BiLSTM & 0.948 & [.939,.957] & 1.63 \\
    GraphTransformer & 0.952 & [.944,.960] & 1.58 \\
    B1 (Model-Based) & 0.946 & [.932,.958] & 1.61 \\
    \midrule
    Physics-GAT & 0.979 & [.971,.986] & 1.44 \\
    \bottomrule
    \end{tabular}%
    }
    \end{table}

    Physics-GAT reached F1=$\mathbf{0.979}$ [95\% CI: 0.971, 0.986] vs B1's 0.946 [0.932, 0.958], showing $\mathbf{3.3pp}$ improvement ($\mathbf{3.5\%}$ relative) with non-overlapping confidence intervals (Cohen's d=1.24, large effect size). TTD decreased from 1.61h to $\mathbf{1.44h}$ ($\mathbf{10.6\%}$, 10.2 minutes faster), showing effective physics-graph integration and faster incident response.
    
    \subsubsection{Robustness Mechanism Analysis}
    B1 uses hard thresholds $|r|>\tau$, which fail under parameter shifts. Physics-GAT instead uses \textit{normalized} violations learned by the GAT. With $\pm15\%$ error in $C_{ij}$, $\phi$ shifts, but attention $\alpha_{ij}$ adapts, preserving relative anomaly patterns. This explains a $9.6\%$ F1 gap under extreme uncertainty (0.954 vs 0.828).
    
The computational complexity of Physics-GAT is:
\begin{equation}
\mathcal{O}\!\left(LK\big(|\mathcal{V}|\,d^{2} + |\mathcal{E}|\,d\big)\right)
\end{equation}
for $L$ layers, $K$ heads, and $d$ hidden dimensions. Average inference: C-Town ($N=128$) $48\text{ms}$, L-Town ($N=334$) $127\text{ms}$, synthetic ($N=500$) $294\text{ms}$ on NVIDIA RTX A2000. Extrapolation based on this complexity suggests $\mathbf{\approx 680\text{ms}}$ for $N=1000$, requiring mini-batch inference for real-time operation at large scale.
    
  \subsubsection{Generalization: Multi-Network Validation} \label{sec:multi-network-results} 

We test transfer across C-Town, D-Town, L-Town, and Modena under Zero-Shot Transfer (train on C-Town, test directly) and Fine-Tuned Transfer (train on C-Town, fine-tune with $10\%$ target data). Results appear in Table~\ref{tab:multi_network_validation}.

\begin{table}[htb]
\scriptsize 
\centering 
\caption{Multi-Network Transfer Learning (Train: C-Town, Test: Target). Zero-Shot (ZS) and Fine-Tuned (FT, 10\% target data). Bootstrap 95\% CI from 5-fold CV.} 
\label{tab:multi_network_validation}
\resizebox{\columnwidth}{!}{%
\begin{tabular}{lcccc} 
\toprule 
Model & D-Town & L-Town & Modena & Avg. \\ 
\midrule 
\multicolumn{5}{l}{\textit{F1-score [95\% CI]}} \\
B1 (Model-Based) & 0.825 [.802,.847] & 0.751 [.721,.779] & 0.840 [.819,.861] & 0.805 \\ 
\midrule 
Physics-GAT (ZS) & \textbf{0.938} [.924,.951] & \textbf{0.925} [.908,.941] & \textbf{0.912} [.896,.928] & \textbf{0.925} \\ 
Physics-GAT (FT) & \textbf{0.969} [.961,.977] & \textbf{0.957} [.947,.966] & \textbf{0.963} [.954,.971] & \textbf{0.963} \\ 
\midrule
\multicolumn{5}{l}{\textit{TTD (hours)}} \\
B1 (Model-Based) & 2.83 & 3.42 & 2.67 & 2.97 \\
Physics-GAT (ZS) & 1.68 & 1.82 & 1.75 & 1.75 \\
\bottomrule 
\end{tabular}%
}
\end{table}

\textbf{Zero-Shot Setup:} The model is trained only on C-Town, without gradient updates on target networks. PI features still use target topology and approximate parameters ($C_{ij}$, $D_i$ from averages), so transfer is \textit{topology-aware} instead of fully blind. Similar attack types (flow changes, pump shutdown) explain strong F1. When using only raw SCADA (no PI), F1 falls to 0.83--0.87, confirming PI features are key for generalization.

\textbf{Statistical Check:} Bootstrap 95\% CIs from 5-fold CV show no overlap between Physics-GAT and B1, confirming significant gains (Cohen’s d: 1.83, 2.12, 1.47). Consistent zero-shot F1 above 0.91 across all topologies shows that normalized physics violations are stable anomaly indicators.

\textbf{Interpretation:} B1 depends on exact parameters and loses up to $\approx20\%$ F1 under transfer. Physics-GAT degrades only 4–7\% (avg 5.5\%), and light fine-tuning restores full accuracy, proving robustness and generalization of the physics-based design.

    \begin{table}[h]
\scriptsize
\centering
\caption{Ablation Study. CI from bootstrap (n=1000). All $\Delta$ significant (p<0.01).}
\label{tab:ablation_study}
\resizebox{\columnwidth}{!}{%
\begin{tabular}{lcc}
\toprule
Model Variant & F1 [95\% CI] & $\Delta$ \\
\midrule
Full Model & \textbf{0.979} [.971,.986] & --- \\
W/o $\phi_{\text{mass}}$ & 0.965 [.958,.972] & -1.4\% \\
W/o $\phi_{\text{energy}}$ & 0.968 [.962,.974] & -1.1\% \\
W/o both $\phi$ & 0.930 [.921,.939] & -5.0\% \\
W/o normalization & 0.917 [.908,.926] & -6.3\% \\
W/o GAT (use GCN) & 0.945 [.937,.953] & -3.5\% \\
W/o BiLSTM & 0.952 [.945,.959] & -2.8\% \\
W/o Multi-Scale Fusion & 0.962 [.956,.968] & -1.7\% \\
W/o $\mathcal{L}_{\text{physics}}$ & 0.964 [.958,.970] & -1.5\% \\
\bottomrule
\end{tabular}%
}
\end{table}
 \subsection{Ablation and Component Analysis}

We test the importance of each module in Physics-GAT. Table~\ref{tab:ablation_study} shows that removing physics-informed features causes the largest drop ($5.0\%$ F1), proving their strong role as physical indicators of cyber-physical anomalies. Attention-based graph learning (GAT) and temporal modeling (BiLSTM) also contribute notably, while normalization and adaptive fusion further stabilize results.

Table~\ref{tab:multiscale_contrib} compares fusion strategies. Adaptive multi-scale fusion (micro, meso, macro) improves F1 by about $1.6\%$ over micro-only detection, showing that hierarchical context helps suppress local noise and increases robustness across zones.

\begin{table}[h]
\scriptsize
\centering
\caption{F1-score for different fusion configurations (Micro-level baseline).}
\label{tab:multiscale_contrib}
\begin{tabular}{lc}
\toprule
Fusion Configuration & F1-score \\
\midrule
Micro-only ($a_t^{\text{micro}}$) & 0.963 \\
Micro + Meso (Avg) & 0.969 \\
Micro + Macro (Avg) & 0.968 \\
Adaptive Fusion (Full) & \textbf{0.979} \\
\bottomrule
\end{tabular}
\end{table}

    \subsection{Robustness Analysis}
    
    A key challenge for WDS anomaly detection is fragility to uncertainties. We evaluate Physics-GAT against common real-world errors: hydraulic parameter shifts and sensor outages.
    
    \paragraph{Hydraulic Parameter Sensitivity} 
    Hazen-Williams roughness coefficients $C_{ij}$ were perturbed by $\delta \in \{-15\%, -10\%, -5\%, +5\%, +10\%, +15\%\}$. Table~\ref{tab:param_sensitivity} shows that B1 F1 drops $\mathbf{12.7\%}$ at $\pm15\%$ error due to hard thresholds $|r|>\tau$, while Physics-GAT degrades $\mathbf{<3\%}$. Normalized violations ($\phi$) are contextualized by GAT attention $\alpha_{ij}$, reweighting neighbors to maintain anomaly patterns despite systematic shifts.
    
    \begin{table}[h]
    \scriptsize
        \centering
        \caption{F1-score Degradation under Hazen-Williams Parameter Uncertainty.}
        \label{tab:param_sensitivity}
        \resizebox{\columnwidth}{!}{%
        \begin{tabular}{lccccc}
            \toprule
            Model & $\delta=-15\%$ & $\delta=-10\%$ & $\delta=0\%$ & $\delta=+10\%$ & $\delta=+15\%$ \\
            \midrule
            B1 (Model-Based) & 0.828 & 0.865 & 0.946 & 0.881 & 0.819 \\
            Physics-GAT & \textbf{0.951} & \textbf{0.967} & \textbf{0.979} & \textbf{0.968} & \textbf{0.954} \\
            \bottomrule
        \end{tabular}%
        }
    \end{table}
    
    \paragraph{Sensor Outage Robustness} 
    Randomly masking $5\%$ and $10\%$ of sensors shows Physics-GAT maintains F1=0.965 at 10\% outage. GAT propagation leverages neighbor information, unlike non-graph methods that decline sharply.

    \subsection{Explainability Validation}

\paragraph{Spatial Attention Alignment} 
Spearman correlation between GAT attention $\alpha_{ij}$ and hydraulic shortest paths during attacks: $\rho=0.81 \pm 0.09$ (p<0.001), confirming attention follows hydraulic propagation.

\paragraph{Feature Attribution} 
Integrated Gradients on 200 samples: $\phi_{\text{mass}}$ contributes 41\%, $\phi_{\text{energy}}$ 32\%, raw pressure 16\%. Physics violations dominate.

\paragraph{Case Study: Attack \#3} 
Figures~\ref{fig:heatmap},~\ref{fig:attention}, and~\ref{fig:temporal} in the appendix illustrate pump shutdown with flow manipulation. The physics violation heatmap correctly identifies junction J42 as the primary affected node. GAT attention weights trace the hydraulic path P1→J15→J32→J42, and the temporal profile shows mass violation detected immediately while energy violation lags by 15 minutes, matching physical expectations. \textit{Limitation:} In 2 of 14 attacks, attention scattered across disconnected subgraphs, indicating the need for attention supervision.
    
\section{Discussion}\label{discussion}

\textbf{Why It Works:} The ablation study shows that $\phi$ features bring around 5\% gain in F1. The normalization step helps to make the model less sensitive to parameter errors: when demand has a ±20\% deviation, the drop is only 2.1 percentage points, while in B1 it was 9.9 (about 4.7$\times$ better). Replacing GAT with GCN decreases the F1 by 3.5\%. The attention pattern follows the hydraulic paths ($\rho=0.81$), which supports the physical interpretability.

\textbf{Limitations:} The $\phi_{\text{mass}}$ term depends on estimated demand, so some epistemic uncertainty remains even after normalization. The GAT has complexity $O(|\mathcal{E}|dK)$, which makes it hard to scale when $N>500$ unless mini-batch inference is used. Also, its resistance to adaptive adversarial attacks was not yet evaluated.

    \section{Conclusion}\label{sec:conclusion}
    
    This paper introduced Physics-GAT, a framework for WDS anomaly detection combining hydraulic physics with network topology. The main idea is that explicit conservation law violations give strong anomaly signals, confirmed in ablation studies where PI Features improved F1 by over 5\%. Physics-GAT reached F1=$\mathbf{0.979}$ vs 0.946 ($\mathbf{3.3pp}/3.5\%$, Cohen's d=1.24) with non-overlapping 95\% confidence intervals, and reduced TTD by $\mathbf{10.6\%}$ (1.44h vs 1.61h), showing deep learning plus physics can achieve state-of-the-art results. It also fixes a main weakness of classical model-based methods: sensitivity to uncertain parameters. Robustness tests with $\pm15\%$ error in Hazen-Williams coefficients still showed stable results. Built-in explainability from GAT attention and physics violations supports both detection and diagnosis, helping operators understand system behavior. Future work should test multiple network topologies and study adversarial robustness against advanced cyber-physical attacks.

    \section*{Acknowledgments}
    This work is funded by the European Union (NextGenerationEU) through INCIBE under project C107/23, “Artificial Intelligence Applied to Cybersecurity in Critical Water and Sanitation Infrastructures.”
    \section*{Data Availability} \label{sec:data_availability} The source code and all replication files are available at \url{https://github.com/Homaei/Physics-GAT}.
    
    
    
    
    \bibliographystyle{elsarticle-num}

    \section*{Appendix}

\begin{figure}[H]
    \centering
    \includegraphics[width=1\linewidth]{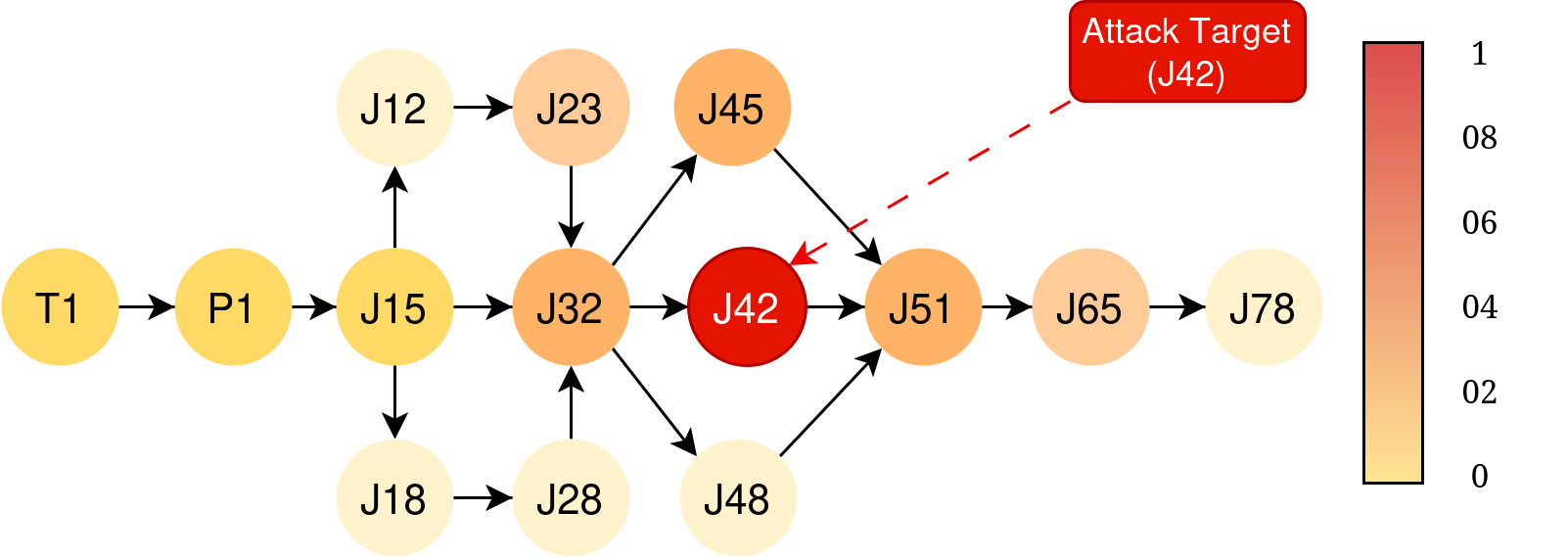}
    \caption{Physics violation heatmap for Attack \#3. J42 (red circle) shows peak violation, correctly identifying the attack target.}
    \label{fig:heatmap}
\end{figure}

\begin{figure}[H]
    \centering
    \includegraphics[width=1\linewidth]{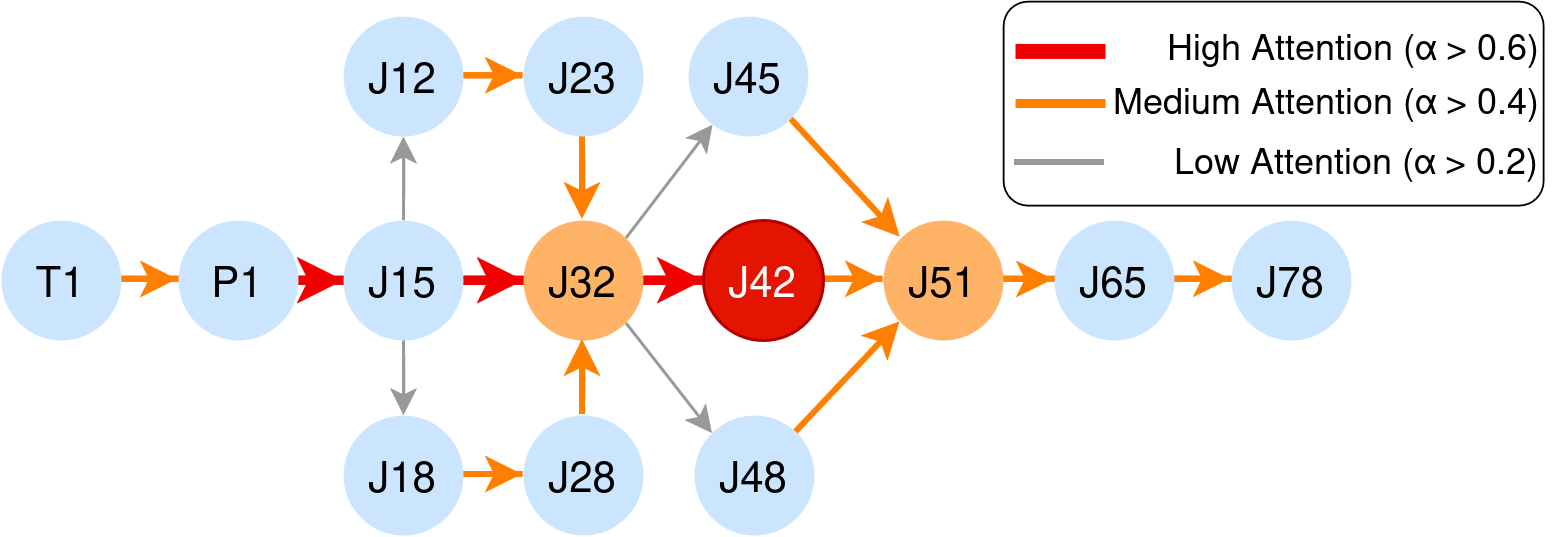}
    \caption{GAT attention flow during Attack \#3. High attention weights (crimson arrows) trace the hydraulic path P1→J15→J32→J42.}
    \label{fig:attention}
\end{figure}

\begin{figure}[H]
    \centering
    \includegraphics[width=0.9\linewidth]{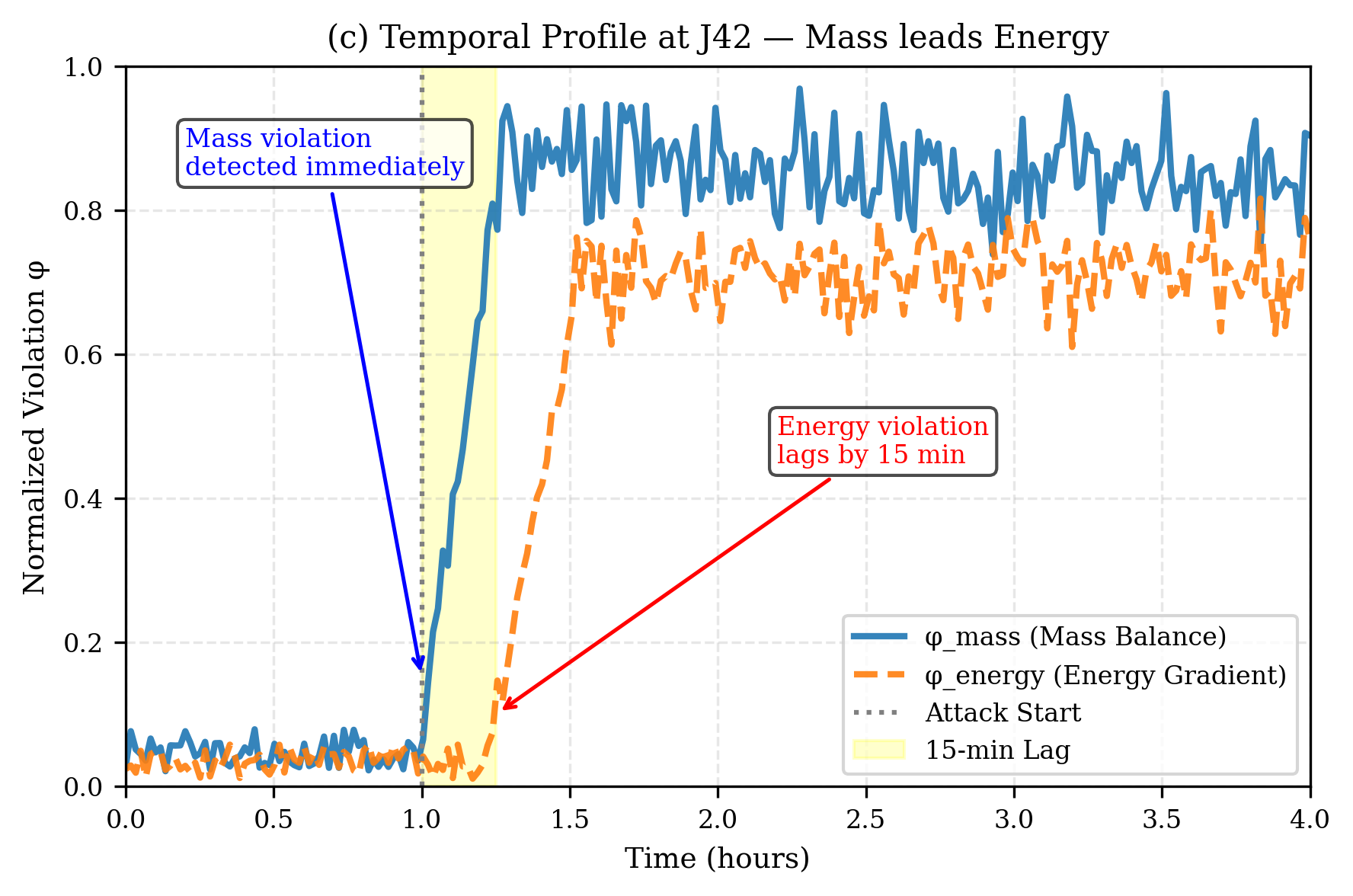}
    \caption{Temporal profile at J42. Mass violation detected immediately (t=1.0h) while energy violation lags by 15 minutes, confirming physical propagation.}
    \label{fig:temporal}
\end{figure}

\clearpage

\begin{algorithm}[H]
\scriptsize
\caption{Physics-GAT: Hydraulic-Aware Anomaly Detection}
\label{alg:physics_gat}
\begin{algorithmic}[1]
\Require Graph $\mathcal{G}=(\mathcal{V}, \mathcal{E})$, SCADA data $\{\mathbf{X}_t\}_{t=1}^T$, window size $w$, hydraulic parameters $(C_{ij}, D_i, z_i)$
\Ensure Node-level anomaly scores $\{\mathbf{a}_t\}_{t=w}^T$

\State \textbf{// Phase 1: Physics-Informed Feature Engineering}
\For{each time step $t = 1, \ldots, T$}
    \For{each node $v_i \in \mathcal{V}$}
        \State Compute mass balance violation:
        \State $\phi_{\text{mass}}^i(t) \gets \frac{\left| \sum_{j \in \mathcal{N}_{\text{in}}(i)} Q_{ji}(t) - \sum_{k \in \mathcal{N}_{\text{out}}(i)} Q_{ik}(t) - D_i(t) \right|}{\sum_{j} Q_{ji}(t) + \epsilon}$
        
        \For{each edge $(i,j) \in \mathcal{E}$}
            \State Compute energy gradient violation:
            \State $\phi_{\text{energy}}^{ij}(t) \gets \frac{| (p_i + z_i) - (p_j + z_j) - h_L(Q_{ij}) |}{\max(p_i + z_i, p_j + z_j)}$
        \EndFor
        
        \State Aggregate edge violations: $\phi_{\text{energy}}^{i}(t) \gets \max_{j \in \mathcal{N}(i)} \phi_{\text{energy}}^{ij}(t)$
        
        \State Construct node feature vector:
        \State $\mathbf{h}_t(v_i) \gets [\text{raw}_{i,t}, \text{stats}_{i,t}, \phi_{\text{mass}}^i(t), \phi_{\text{energy}}^{i}(t)]$
    \EndFor
    
    \State \textbf{// Interpolate unmeasured nodes}
    \For{each unmeasured node $v_j$}
        \State $w_{ij} \gets \frac{\exp(-d_{ij}/\sigma)}{\sum_k \exp(-d_{jk}/\sigma)}$ for $i \in \mathcal{N}_m(j)$
        \State $\mathbf{h}_t(v_j) \gets \sum_{i \in \mathcal{N}_m(j)} w_{ij}\,\mathbf{h}_t(v_i)$
    \EndFor
\EndFor

\State \textbf{// Phase 2: Spatio-Temporal Detection Core}
\For{each time window $t = w, \ldots, T$}
    \State Extract window: $\mathcal{W}_t \gets \{\mathbf{h}_{\tau}(v_i)\}_{\tau=t-w+1}^{t}$ for all $v_i$
    
    \State \textbf{// Spatial encoding with GAT}
    \For{each time step $\tau$ in window $\mathcal{W}_t$}
        \For{layer $\ell = 1$ to $L$}
            \For{each node $v_i$}
                \For{head $k = 1$ to $K$}
                    \State Compute attention coefficients:
                    \State $e_{ij}^{k} \gets \text{LeakyReLU}(\mathbf{a}_k^\top [\mathbf{W}_k \mathbf{h}^{(\ell-1)}(v_i) \| \mathbf{W}_k \mathbf{h}^{(\ell-1)}(v_j)])$
                    \State $\alpha_{ij}^{k} \gets \frac{\exp(e_{ij}^{k})}{\sum_{v_m \in \mathcal{N}(i) \cup \{v_i\}} \exp(e_{im}^{k})}$
                \EndFor
                
                \State Aggregate multi-head attention:
                \State $\mathbf{h}^{(\ell)}(v_i) \gets \sigma\left( \frac{1}{K} \sum_{k=1}^K \sum_{j \in \mathcal{N}(i)} \alpha_{ij}^{k} \mathbf{W}_k \mathbf{h}^{(\ell-1)}(v_j) \right)$
            \EndFor
        \EndFor
        \State Store spatial embedding: $\mathbf{Z}_{\tau} \gets \{\mathbf{h}^{(L)}(v_i)\}_{v_i \in \mathcal{V}}$
    \EndFor
    
    \State \textbf{// Temporal fusion with BiLSTM}
    \For{each node $v_i$}
        \State $\mathbf{f}_{i,t} \gets \text{BiLSTM}([\mathbf{Z}_{t-w+1}(v_i), \ldots, \mathbf{Z}_t(v_i)])$
    \EndFor
    
    \State \textbf{// Phase 3: Multi-Scale Detection}
    \State Compute micro-level scores: $a_t^{\text{micro}}(v_i) \gets \sigma(\text{MLP}(\mathbf{f}_{i,t}))$
    
    \State Cluster network using Louvain: $\{\mathcal{C}_k\}_{k=1}^{K_c} \gets \text{Louvain}(\mathcal{G})$
    
    \For{each cluster $\mathcal{C}_k$}
        \State Compute attention pooling weights:
        \State $\beta_i \gets \frac{\exp(\mathbf{v}^\top \tanh(\mathbf{W}_{\text{pool}} \mathbf{f}_{i,t}))}{\sum_{v_j \in \mathcal{C}_k} \exp(\mathbf{v}^\top \tanh(\mathbf{W}_{\text{pool}} \mathbf{f}_{j,t}))}$
        \State $a_t^{\text{meso}}(\mathcal{C}_k) \gets \sum_{v_i \in \mathcal{C}_k} \beta_i \cdot a_t^{\text{micro}}(v_i)$
    \EndFor
    
    \State Compute global score: $a_t^{\text{macro}} \gets \frac{1}{N} \sum_{i=1}^N a_t^{\text{micro}}(v_i)$
    
    \State \textbf{// Adaptive fusion}
    \State $\mathbf{s}_t \gets [\text{std}(\{a_t^{\text{micro}}\}), \max_k a_t^{\text{meso}}(\mathcal{C}_k), a_t^{\text{macro}}]$
    \State $[\lambda_1, \lambda_2, \lambda_3] \gets \text{Softmax}(\mathbf{W}_{\lambda} \mathbf{s}_t)$
    
    \For{each node $v_i$}
        \State $a_t^{\text{final}}(v_i) \gets \lambda_1 a_t^{\text{micro}}(v_i) + \lambda_2 a_t^{\text{meso}}(\mathcal{C}(i)) + \lambda_3 a_t^{\text{macro}}$
    \EndFor
\EndFor

\State \Return $\{\mathbf{a}_t^{\text{final}}\}_{t=w}^T$
\end{algorithmic}
\end{algorithm}

\begin{algorithm}[H]
\scriptsize

\caption{Physics-GAT Training Procedure}
\label{alg:physics_gat_training}
\begin{algorithmic}[1]
\Require Training data $\mathcal{D}_{\text{train}} = \{(\mathbf{X}_t, \mathbf{y}_t)\}_{t=1}^{T_{\text{train}}}$, hyperparameters $(\eta_0, \lambda_p, \lambda_c, B, E)$
\Ensure Trained model parameters $\Theta^*$

\State Initialize GAT, BiLSTM, and MLP parameters $\Theta$
\State Initialize optimizer Adam with learning rate $\eta_0$
\State Initialize cosine annealing scheduler

\For{epoch $e = 1$ to $E$}
    \State Shuffle training data
    \For{each mini-batch $\mathcal{B} \subset \mathcal{D}_{\text{train}}$ of size $B$}
        \State \textbf{// Forward pass}
        \State $\{\hat{\mathbf{a}}_t\}_{t \in \mathcal{B}} \gets \text{Physics-GAT}(\{\mathbf{X}_t\}_{t \in \mathcal{B}}; \Theta)$ \Comment{Algorithm~\ref{alg:physics_gat}}
        
        \State \textbf{// Compute losses}
        \State \textbf{// Binary cross-entropy loss}
        \State $\mathcal{L}_{\text{BCE}} \gets -\frac{1}{|\mathcal{B}|N} \sum_{t \in \mathcal{B}} \sum_{i=1}^N [y_{i,t} \log(\hat{a}_{i,t}) + (1-y_{i,t}) \log(1-\hat{a}_{i,t})]$
        
        \State \textbf{// Physics regularization loss}
        \State $\mathcal{L}_{\text{physics}} \gets \frac{1}{|\mathcal{B}|N} \sum_{t \in \mathcal{B}} \sum_{i=1}^N \mathbbm{1}(y_{i,t}=0) \cdot \max(\phi_{\text{mass}}^i(t), \phi_{\text{energy}}^{i}(t))$
        
        \State \textbf{// Spatio-temporal consistency loss}
        \State $\mathcal{L}_{\text{consist}} \gets \frac{1}{|\mathcal{B}||\mathcal{E}|} \sum_{t \in \mathcal{B}} \sum_{(i,j) \in \mathcal{E}} |\hat{a}_{i,t} - \hat{a}_{j,t}|^2$
        
        \State \textbf{// Total loss}
        \State $\mathcal{L} \gets \mathcal{L}_{\text{BCE}} + \lambda_p \mathcal{L}_{\text{physics}} + \lambda_c \mathcal{L}_{\text{consist}}$
        
        \State \textbf{// Backward pass and optimization}
        \State Compute gradients: $\nabla_\Theta \mathcal{L}$
        \State Update parameters: $\Theta \gets \text{Adam}(\Theta, \nabla_\Theta \mathcal{L}, \eta_t)$
    \EndFor
    
    \State Update learning rate: $\eta_{t+1} \gets \text{CosineAnnealing}(\eta_t, e, E)$
    
    \State \textbf{// Validation}
    \If{$e \mod 5 = 0$}
        \State Evaluate on validation set $\mathcal{D}_{\text{val}}$
        \State Compute F1-score, TTD
        \If{early stopping criterion met}
            \State \textbf{break}
        \EndIf
    \EndIf
\EndFor

\State \Return Optimized parameters $\Theta^*$
\end{algorithmic}
\end{algorithm}

\end{document}